\documentclass[letterpaper]{article} % DO NOT CHANGE THIS
\usepackage{aaai2026}  % DO NOT CHANGE THIS
\usepackage{times}  % DO NOT CHANGE THIS
\usepackage{helvet}  % DO NOT CHANGE THIS
\usepackage{courier}  % DO NOT CHANGE THIS
\usepackage[hyphens]{url}  % DO NOT CHANGE THIS
\usepackage{graphicx} % DO NOT CHANGE THIS
\usepackage{natbib}  % DO NOT CHANGE THIS AND DO NOT ADD ANY OPTIONS TO IT
\usepackage{caption} % DO NOT CHANGE THIS AND DO NOT ADD ANY OPTIONS TO IT
\usepackage{algorithm}
\usepackage{algorithmic}

 \usepackage{listings}
\usepackage{array}
\usepackage{subcaption}

\usepackage{newfloat}
\usepackage{listings}
\DeclareCaptionStyle{ruled}{labelfont=normalfont,labelsep=colon,strut=off} % DO NOT CHANGE THIS
\floatstyle{ruled}
\newfloat{listing}{tb}{lst}{}
\floatname{listing}{Listing}
\title{Large Language Models'\\ Internal Perception of Symbolic Music}

\author {
   Andrew Shin\textsuperscript{\rm 1},
    Kunitake Kaneko\textsuperscript{\rm 1}
}
\affiliations {
    \textsuperscript{\rm 1}Keio University\\
    shin@inl.ics.keio.ac.jp, kankeo@inl.ics.keio.ac.jp
}
\usepackage{bibentry}
\begin{document}

\maketitle

\begin{abstract}
Large language models (LLMs) excel at modeling relationships between strings in natural language and have shown promise in extending to other symbolic domains like coding or mathematics. However, the extent to which they implicitly model symbolic music remains underexplored. This paper investigates how LLMs represent musical concepts by generating symbolic music data from textual prompts describing combinations of genres and styles, and evaluating their utility through recognition and generation tasks. We produce a dataset of LLM-generated MIDI files without relying on explicit musical training. We then train neural networks entirely on this LLM-generated MIDI dataset and perform genre and style classification as well as melody completion, benchmarking their performance against established models. Our results demonstrate that LLMs can infer rudimentary musical structures and temporal relationships from text, highlighting both their potential to implicitly encode musical patterns and their limitations due to a lack of explicit musical context, shedding light on their generative capabilities for symbolic music.
\end{abstract}


\section{Introduction}
\label{sec:introduction}
Large language models (LLMs), trained exclusively on vast corpora of text strings, have demonstrated remarkable proficiency in capturing not only linguistic structures but also intricate aspects of other symbolic domains \cite{Pavlick2023SymbolsAG} embedded within those strings . This capability stems from their ability to model relationships between sequences of characters, enabling them to infer and generate representations far beyond mere syntax. For instance, LLMs can produce executable code in programming languages, effectively simulating computational logic and algorithmic processes that mirror real-world problem-solving \cite{Wang2024ExecutableCA}. Similarly, they can solve mathematical problems, such as algebraic equations or geometric proofs, by leveraging textual descriptions of logical relationships and formal rules \cite{Shao2024DeepSeekMathPT, Yang2024Qwen25MathTR}. These examples illustrate that LLMs, despite their string-based training, can implicitly encode structured knowledge across diverse domains, raising intriguing questions about the breadth and depth of their world knowledge.

In contrast to their well-documented success in code generation and mathematics, the implicit modeling of symbolic music within LLMs remains relatively unclear and underexplored. Music, as a symbolic system, combines hierarchical structures and temporal dynamics in ways analogous to language, yet its abstract, non-linguistic nature poses unique challenges. While LLMs can generate textual descriptions of music or even lyrics, their ability to directly produce structured musical data, such as MIDI sequences encoding pitch, duration, and velocity, has received little attention. This gap is significant as music offers a rich testbed for probing how LLMs generalize pattern recognition to domains beyond natural language, potentially revealing whether their string-modeling prowess extends to the temporal and harmonic relationships that define the musical world.

In this regard, \cite{Sharma2024AVC} provides an interesting foundation, where they investigated what modeling relationships between strings teaches LLMs about the visual world. They systematically evaluated LLMs’ abilities to generate and recognize visual concepts by representing images as code, circumventing the models’ inability to process pixels directly. Their findings that precise string modeling enables LLMs to encode aspects of visual complexity and even support self-supervised visual representation learning suggest that LLMs can transcend their textual origins to grasp structured, non-linguistic domains. This insight motivates our exploration of music, a domain similarly abstract and structured, where we hypothesize that LLMs might learn musical concepts involving genre, style, and melody through analogous text-to-symbol mappings, offering a parallel lens into their representational capabilities.

It is important to acknowledge that the term "understanding" when applied to Large Language Models is a complex and highly debated topic within the AI community. This paper does not claim that LLMs possess human-like "understanding" of music. Rather, we use terms like "perceive," "infer," or "model" to refer to the LLM's capacity to identify and generate structured patterns in symbolic music based on its vast textual pre-training. We recognize that this is distinct from cognitive understanding, which is a nuanced and multi-faceted human capability. \cite{Mitchell2022TheDO} and \cite{Jacobs2019MeasurementAF} highlight the challenges in defining and reliably measuring "understanding" in AI, and demonstrates specific failure modes of LLMs that suggest a lack of human-like comprehension. Our aim is to investigate the computational phenomenon of LLMs processing symbolic music, not to make claims about their cognitive state. The evidence presented herein focuses on the LLM's ability to process and generate structured sequences, which, while fascinating, is fundamentally different from a human's rich, experiential understanding of music.

In this paper, we systematically assess how LLMs model the musical world by performing symbolic music recognition and generation through LLM. In particular, we set out to investigate two questions; 1) To what extent do LLMs implicitly encode musical information from their text-only pre-training, enabling them to "perceive," i.e. process and generate structured symbolic music, from textual prompts? 2) Can LLM-generated symbolic music data be effectively used to train neural networks for music classification and generation tasks? As a by-product, we generate a dataset of MIDI files from textual prompts specifying combinations of genres and styles. Each song is generated without predefined musical templates, relying entirely on the LLM’s interpretation of prompt relationships. Furthermore, we train a convolutional neural network on this dataset to classify genres and styles, evaluating the distinctiveness of LLM-generated musical features, and additionally adapt the model into a transformer-based decoder to perform melody completion task. Our approach tests the hypothesis that string modeling equips LLMs with the ability to implicitly derive rudimentary musical structures, expressed through symbolic outputs rather than explicit musical training.

The implications of this work are two-fold: it advances our understanding of LLMs as general-purpose pattern learners capable of cross-domain synthesis, and it opens new avenues for generative music systems powered by text-driven AI. By demonstrating that LLMs can generate classifiable musical structures and plausible melodies from prompts alone, we highlight their potential to implicitly derive a "musical world" representation akin to their grasp of code or visual concepts. This suggests that the power of string modeling lies in its ability to abstract and generalize across symbolic systems, a finding with significant ramifications for AI creativity and interdisciplinary applications in music informatics. %Our study thus contributes to the broader discourse on how LLMs encode world knowledge, positioning music as a critical frontier in this exploration.

\section{Related Work}
The investigation into how large language models (LLMs) extrapolate knowledge from string modeling to non-linguistic domains has gained traction with \cite{Sharma2024AVC} providing a foundational exploration in the visual domain that directly inspires our work. They systematically evaluate LLMs’ capacity to generate and recognize visual concepts by representing images as code, bypassing the models’ inability to process raw pixels. Their results demonstrate that precise string modeling enables LLMs to encode complex visual features, such as shapes and spatial relationships, and even support self-supervised visual representation learning. This paradigm of using text as a bridge to structured symbolic outputs parallels our approach, where we leverage LLM to generate MIDI files from textual prompts describing musical genres and styles. While \cite{Sharma2024AVC} focus on static visual representations, our work extends this concept to the temporal and hierarchical domain of music, probing whether LLMs can similarly capture the dynamic patterns of melody, harmony, and dynamics through string relationships.

There have been notable previous works that examine LLMs in the context of symbolic music. Text2MIDI \cite{Bhandari2024Text2midiGS} introduces an end-to-end model for generating MIDI files from textual descriptions using an LLM encoder paired with an autoregressive transformer decoder. They capitalize on LLMs’ text-processing strengths to produce controllable MIDI sequences reflecting music theory terms, such as chords, keys, tempo, and validate through automated and human evaluations. While effective, Text2MIDI relies on a specialized encoder-decoder architecture tailored for music generation, contrasting with our approach of using a single, unmodified LLM to directly output strings to be converted to MIDI data. As such, our method avoids additional architectural complexity, testing the raw generative capacity of a general-purpose LLM. $\mathrm{M}^6(\mathrm{GPT})^3$ \cite{Pocwiardowski2024M6GPT3GM} further explores text-driven music generation, employing an autoregressive transformer to map natural language prompts to JSON composition parameters, followed by a genetic algorithm for melody generation and probabilistic methods  for percussion. This hybrid approach generates multi-minute MIDI compositions with complex structures, outperforming neural baselines on musically meaningful metrics. Unlike $\mathrm{M}^6(\mathrm{GPT})^3$, which integrates rule-based algorithms with LLMs to enforce musical coherence, our work relies entirely on the LLM’s output, avoiding post-processing or external algorithms, enabling us to directly assess the LLM’s implicit pattern recognition abilities for musical structure. This advantage lies in its ability to directly probe the LLM's inherent capabilities without confounding factors introduced by domain-specific training or rule-based post-processing. While specialized models can achieve superior performance, understanding the baseline capabilities of general-purpose LLMs in novel domains is intrinsically valuable for advancing the broader field of AI. It helps us to discern what knowledge is genuinely learned from massive text corpora and what requires explicit domain adaptation.

MuPT \cite{Qu2024MuPTAG} investigates LLMs in music pre-training, arguing that ABC Notation aligns better with LLM strengths than MIDI, and proposes Synchronized Multi-Track ABC Notation (SMT-ABC) to maintain coherence across tracks. Pre-trained on extensive symbolic music data, MuPT’s models handle up to 8192 tokens and explore scaling laws for performance. While MuPT leverages ABC Notation’s text-like structure for pre-training, our work uses MIDI directly, aligning with its prevalence in symbolic music tasks and avoiding the need for notation conversion. NotaGen \cite{Wang2025NotaGenAM} introduces a symbolic music generation model for classical sheet music, pre-trained on 1.6M ABC Notation pieces, fine-tuned on 9K classical compositions and enhanced via reinforcement learning with the CLaMP-DPO method. Subjective tests show NotaGen outperforming baselines against human compositions. While NotaGen excels in classical music aesthetics through extensive pre-training and fine-tuning, our work targets a broader genre-style spectrum without pre-training, relying solely on prompt-driven generation. %This makes our approach more flexible and less resource-intensive, as we avoid large-scale musical corpora and complex training paradigms. Furthermore, our dual focus on classification (13 genres, 25 styles) and melody generation offers a more comprehensive evaluation of LLM musicality compared to NotaGen’s classical-specific scope.

Text-to-symbolic-music generation has 
also seen notable progress. MuseCoco \cite{Lu2023MuseCocoGS} uses musical attributes as a bridge between text descriptions and symbolic music generation, demonstrating superior performance compared to GPT-4 in musicality and controllability. MeloTrans \cite{Wang2024MeloTransAT} employs principles of motif development rules and outperforms existing music generation models. XMusic \cite{Tian2025XMusicTA} emerged as a generalized symbolic music generation framework supporting various input modalities, including text, and focusing on emotional control and high-quality outputs. While these models represent significant progress in combining natural language processing with music composition techniques, their models require a large-scale pre-training with symbolic music and text data, and fundamentally deviate from the objective of our work of examining LLMs trained with text only.

In short, our work stands out by using a single, unmodified LLM to generate a diverse, large-scale MIDI dataset without external rules or pre-training, directly testing the limits of string modeling in music. We integrate classification and generation tasks, providing a holistic assessment of LLM-derived musical representations. While the abstract noted that "internal perception of symbolic music remains underexplored," it is important to clarify that this refers to the specific direct probing of general-purpose, text-only LLMs without musical fine-tuning, as opposed to the broader field of LLMs applied to music, where significant work exists. %This distinction is crucial for understanding the unique contribution of our study.%This purity and breadth distinguish our contribution, shedding light on LLMs’ raw capacity to encode the musical world through text alone.

\section{Data Generation}\label{sec:data_generation}
\begin{figure*}[t!]
    \begin{subfigure}[b]{0.98\textwidth}
        \includegraphics[width=\textwidth]{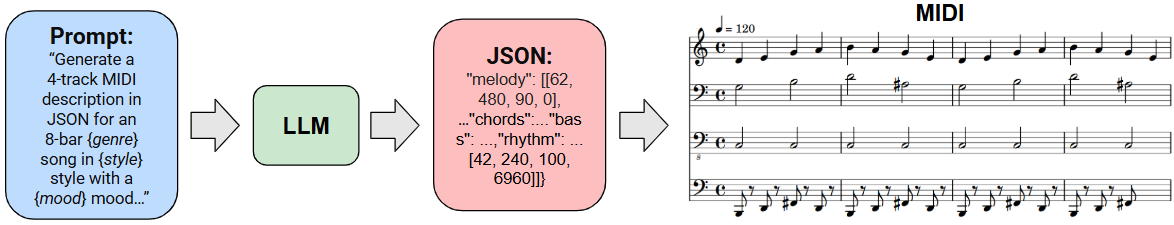}
        \caption{Generation of MIDI file using LLM. Prompt asks LLM to generate a JSON description of MIDI, specifying genre, style, and mood. The resulting JSON output is subsequently converted to an actual MIDI file.}
        \label{fig:gen}
        \hfill
        \vspace{1mm}
            \end{subfigure}
    \begin{subfigure}[b]{0.98\textwidth}
        \includegraphics[width=\textwidth]{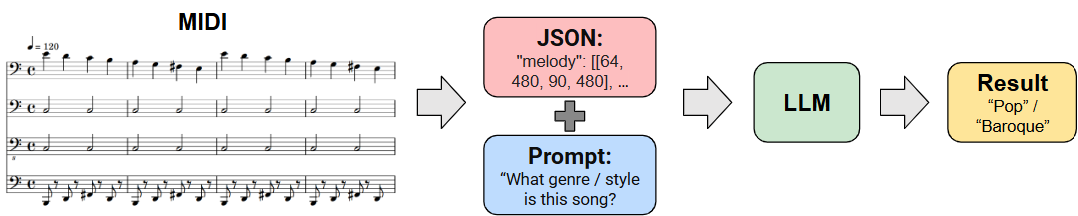}
        \caption{Recognition of MIDI file using LLM. Existing MIDI file is converted to JSON format and, along with a prompt asking for the its genre or style, is input to LLM which returns the classification result.}
        \label{fig:recog}        \hfill
        \vspace{1mm}

    \end{subfigure}
    
    %\vspace{1em}
    
    \begin{subfigure}[b]{0.98\textwidth}
        \includegraphics[width=\textwidth]{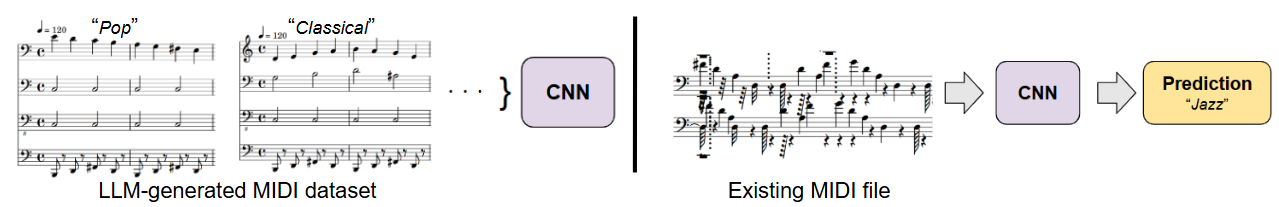}
        \caption{We train a CNN with our LLM-MIDI dataset (left), and perform inference on existing MIDI files to perform genre and style classifications (right).}
        \label{fig:train}        \hfill
        \vspace{1mm}

    \end{subfigure}
    \begin{subfigure}[b]{0.98\textwidth}
        \includegraphics[width=\textwidth]{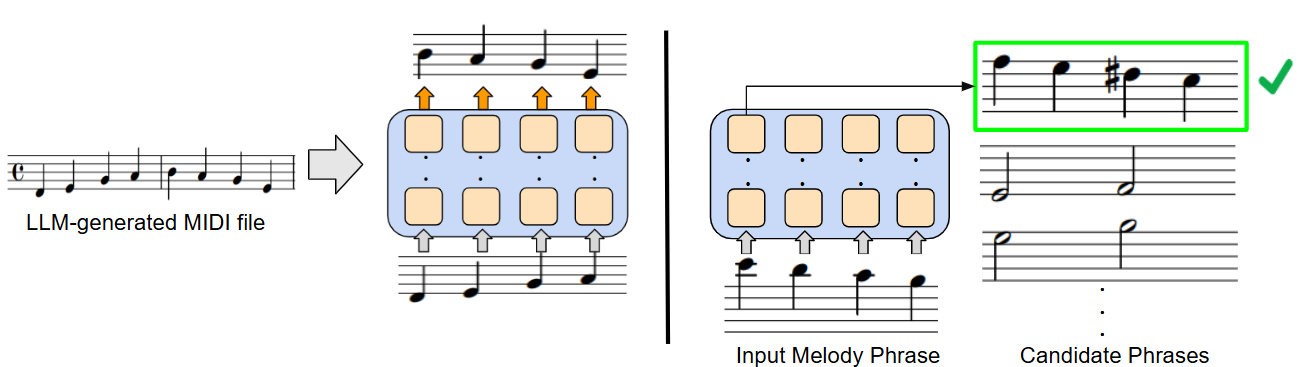}
        \caption{We train a transformer with our LLM-MIDI dataset to predict the next melodic phrase (left), and find the most likely candidate given the input melodic phrase in evaluation (right).}
        \label{fig:melody}        \hfill
        \vspace{1mm}

    \end{subfigure}
    
    \caption{Illustrations of the workflow of our experiments.}
    \label{fig:experiments}
\end{figure*}
To investigate LLMs' impliciit modeling of symbolic music, we generated a novel LLM-generated dataset of MIDI files, spanning 13 genres from the TOP-MAGD taxonomy and 25 styles from the MASD framework \cite{Ferraro2018OnLG}. %This section details the methodology employed to create this dataset, emphasizing the text-to-MIDI generation process, prompt design, and error handling strategy.

\begin{figure*}[t!]
    \begin{subfigure}[b]{0.50\textwidth}
        \includegraphics[width=\textwidth]{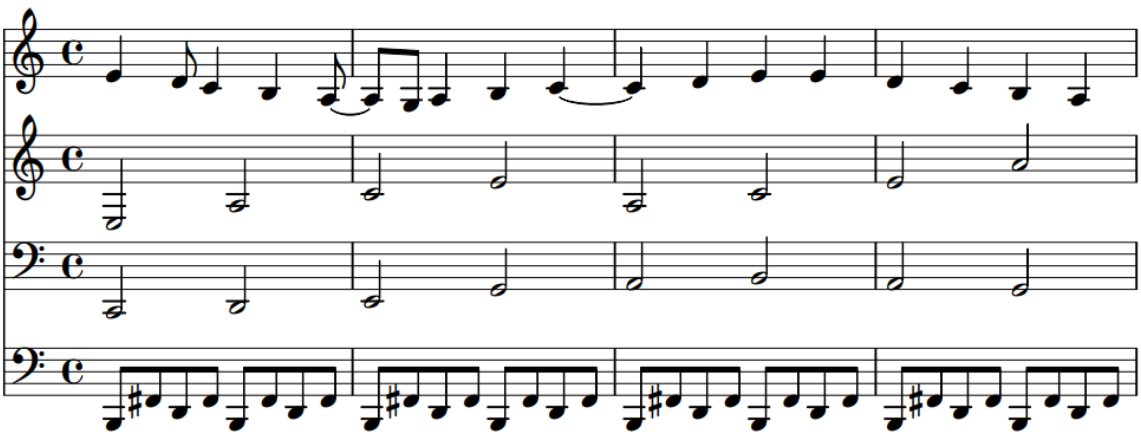}
        \caption{MIDI sample generated for \textit{pop} genre in \textit{romantic} style.}
        \label{fig:ex1}

    \end{subfigure}
    \hfill
    \begin{subfigure}[b]{0.47\textwidth}
        \includegraphics[width=\textwidth]{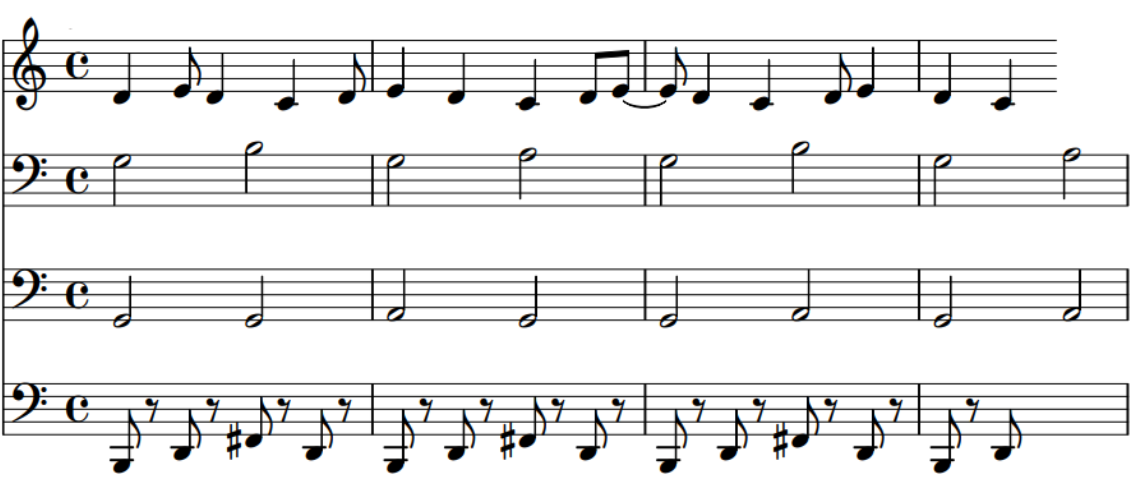}
        \caption{MIDI sample generated for \textit{rock} genre in \textit{minimalist} style.}
        \label{fig:ex2}
    \end{subfigure}
    \caption{Example songs from LLM-MIDI dataset.}
    \label{fig:ex}
\end{figure*}

%\subsection{Prompt Design and MIDI Structure}
We designed textual prompts to elicit four-track MIDI sequences, each comprising melody, chords, bass, and rhythm tracks. The prompts followed the format ``\textit{[genre] song in [style] manner}'', augmented with a randomly selected mood, e.g. '\textit{happy}' or '\textit{sad},' to enhance diversity. We instructed the LLM to output a JSON string encoding 8-bar sequences, with each track defined as a list of tuples \texttt{(pitch, duration, velocity, start\_time)}. We also imposed specific constraints; pitch values range from 0 to 127, duration values are 240 (eighth note), 480 (quarter note), or 960 (half note) ticks, velocity ranges from 0 to 127, and start times span 0 to 7680 ticks. For the rhythm track, drum pitches were restricted to 35 (kick), 38 (snare), and 42 (hi-hat), aligning with MIDI percussion standards. The instruction explicitly demanded a pure JSON string as output, which was then parsed and validated to ensure all four tracks were present and values conformed to the specified ranges. The generation process iterated over all 325 genre-style combinations, producing 50 songs per combination, summing up to  generation of 16,250 files. This amounts to 1250 songs per class for genre classification and 650 songs per class for style classification. We adopted temperature parameter varying from 0.6 to 1.0 (incremented by 0.04 per song index modulo 10) to introduce variability, and set the maximum number of tokens to 1200 to accommodate complete JSON outputs, addressing truncation issues observed in preliminary tests. Successful outputs were converted to MIDI files using the \texttt{mido} library, with each track assigned appropriate instruments (piano for melody and chords, electric bass for bass, and MIDI channel 10 for rhythm). Our generated dataset consists of 780k note events, which required 7.3M tokens. We used GPT-4 as our LLM, although any other LLM can used. Figure~\ref{fig:gen} describes the generation process. Table~\ref{tab:dataset_chars} summarizes the statistics of this dataset.

\begin{table}[t!]
\centering
\small
\caption{statistics of the LLM-MIDI dataset.}
\begin{tabular}{c|c}
\hline
\bf Characteristic & \bf Description \\ \hline
\textbf{Total \# Files} & 16,250 \\
\textbf{\# of Genre Classes} & 13 (1250 per class) \\
\textbf{\# of Style Classes} & 25 '(650 per class)\\
%\textbf{Songs per Class} & 1250 for genre, 650 for style\\
\textbf{Tracks per File} & 4 (melody, chords, bass, rhythm) \\
\textbf{Sequence Length} & 8 bars (7680 ticks at 128 time steps)\\
\textbf{Pitch Range} & 0–127; For rhythm, 35/38/42\\
\textbf{Duration Values} & 240/480/960 \\
\textbf{Velocity Range} & 0–127\\
\textbf{Generation Model} & GPT-4 \\
%\textbf{Prompt Format} & ``[genre] song in [style] manner'' with random mood (happy, sad, energetic, calm, mysterious) \\
\textbf{Temperature} & 0.6–1.0 \\
\textbf{Max Tokens} & 1200  \\
\textbf{Total Note Events} & 783,928 \\
\textbf{Mood Variation} & happy/sad/energetic/calm/mysterious \\
%\textbf{Note Density per Track} & Melody: ~16 events, Chords: ~8, Bass: ~8, Rhythm: ~16 (targeted by prompt, with LLM-induced variation) \\
%\textbf{Pitch Usage Diversity} & Melody: Typically 5–15 unique pitches per song; Chords: 3–6 per progression; Bass: 1–3; Rhythm: 3 fixed \\
%\textbf{Chord Progression Diversity} & Varies within class (e.g., I-IV-V-I common in Pop, more complex in Jazz), driven by LLM interpretation \\
%\textbf{Rhythmic Diversity} & Rhythm track varies in kick/snare/hi-hat patterns (240–960 tick spacing), reflecting style differences \\
%\textbf{Uniqueness Metric} & Estimated 80–90\% unique songs per class (based on temperature range and mood randomization) \\
\hline
\end{tabular}
\label{tab:dataset_chars}
\end{table}

%The resulting dataset, stored in \texttt{midi\_dataset/}, comprises up to 16,250 MIDI files organized into 325 subdirectories (e.g., \texttt{pop\_baroque/001.mid}). Each file represents an 8-bar composition with four tracks, totaling approximately 780,000 note events across the dataset (assuming ~48 events per file). The generation process, executed on a standard CPU with internet access, took approximately 4.5–9 hours, with a cost of \$8.44–\$10.13 based on OpenAI API pricing (\$0.0005/1K input tokens, \$0.0015/1K output tokens) for an estimated 7.3 million tokens. Skipped songs, if any, were minimal due to the increased \texttt{max\_tokens} and retry strategy, ensuring a near-complete dataset for subsequent analysis.

This data generation approach leverages the raw generative capacity of LLM for symbolic music, avoiding musical templates or pre-training, to directly test the LLM’s ability to model musical relationships from text prompts alone. In the rest of the paper, we refer to this dataset as LLM-MIDI. Figure~\ref{fig:ex} shows sheet music for example songs from LLM-MIDI dataset.

It is important to acknowledge several methodological limitations imposed by these constraints. The 8-bar length, fixed 4/4 time signature, and limited rhythmic values severely restrict the complexity of musical structures that can be generated. Furthermore, the specific instrumentation and the restriction to only 3 drum instruments, along with the fixed tempo of 120 BPM, are deeply incompatible with the stylistic diversity expected from some genres and styles within the TOP-MAGD and MASD taxonomies. For instance, genres like free jazz or complex electronic music often feature irregular meters, highly varied rhythmic patterns, and diverse instrumentation that cannot be adequately captured by these constraints. Similarly, not all mood vocabulary pairings are musically compatible with all genre/style combinations. These artificial and sometimes musically inappropriate constraints inherently limit the expressive range of the LLM-MIDI dataset, potentially leading to a lower upper bound on performance in downstream tasks and obscuring the LLM's full generative potential. These design choices were made to simplify the initial controlled experiment and ensure consistent JSON output for parsing, but their impact on musical realism and complexity should be considered when interpreting the results. Future work will explore generating music with fewer constraints to better reflect the complexity of real musical datasets.

%When classifying music from the existing TOP-MAGD and MASD datasets, these input MIDI files were not subjected to the same musical constraints (e.g., 8-bar length, fixed instrumentation, limited rhythm) as the LLM-generated MIDI. This difference means the LLM-MIDI dataset is systematically simpler and less diverse in its musical structure than the real-world datasets, which can indeed bias the results for the experiments described in Figures 1b and 1c. This distinction is crucial for a fair interpretation of the performance gap between models trained on LLM-MIDI and those trained on human-composed data. Our primary goal was to assess the LLM's inherent ability to generate musically structured data from text, and the subsequent trainability of neural networks on this constrained output, rather than directly comparing generated music with unconstrained human compositions. Future work will explore generating music with fewer constraints to better reflect the complexity of real musical datasets.

\section{Experiments}

We perform genre and style classification and melody completion tasks in various settings. First, we examine LLM's ability to recognize the attributes of existing MIDI files by directly asking LLM to classify the input. Subsequently, using LLM-MIDI dataset generated in Section~\ref{sec:data_generation}, we train a convolutional neural network and see its performance on classification. For melody completion, we train a transformer model with LLM-MIDI dataest.

The experimental protocol primarily focuses on evaluating the capacity of a general-purpose LLM to implicitly encode musical information from its text-only pre-training and to generate symbolic music. It also assesses the utility of this LLM-generated data for training downstream neural networks. While we acknowledge that a comprehensive understanding would ideally involve extensive ablation studies on prompts, music representation, and a broader range of LLM architectures, this paper lays foundational work for this novel direction. Our primary aim is to establish a baseline for LLM-driven symbolic music generation and its utility, which naturally points to these deeper investigations as crucial future work. %We also note that the black-box nature of LLM pre-training data means we cannot confirm specific exposure to MIDI data; our work aims to test the emergent capabilities from general text-based pre-training.

\subsection{Setting}
\label{subsec:setting}
In order to examine musical recognition ability of LLM, we convert MIDI files from TOP-MAGD and MASD dataset's test splits into JSON representation, and feed it as input into LLM along with a prompt that asks to classify its genre or style (Figure~\ref{fig:recog}). This setup tests the LLM’s zero-shot ability to interpret symbolic music directly from its internal representations, bypassing traditional supervised learning on musical data. %The implication is significant: if successful, it suggests LLMs can infer musical genre and style from raw MIDI data alone, leveraging their text-based training to bridge into the musical domain, potentially reducing the need for domain-specific datasets and models in preliminary music analysis tasks.

For supervised model, we designed a simple vanilla CNN with two convolutional layers and a fully connected layer, using separate output heads for genres and styles. The input is a 4-channel piano roll representation with dimensions \(4 \times 128 \times 128\), where 128 time steps approximate 8 bars and 128 pitches cover the MIDI range. The first convolutional layer has 32 filters with a \(3 \times 3\) kernel, while the second layer has 64 filters, followed by \(2 \times 2\) max pooling. The flattened output is processed through a 128-unit fully connected layer with 50\% dropout, yielding logits for 13 genres or 25 styles via linear heads. This design leverages spatial and temporal patterns across tracks, suitable for symbolic music classification. We trained this CNN with LLM-MIDI dataset and use TOP-MAGD and MASD datasets for evaluation. We measured performance using weighted F1-score. Figure~\ref{fig:train} illustrates our workflow for classification task.

\begin{table}[t!]
\centering
\caption{Results on genre and style classification tasks. LLM indicates directly inputting JSON format of MIDI file along wtih prompt into LLM, whereas CNN + LLM-MIDI indicates a CNN model trained with LLM-MIDI dataset. Note that all baseline models are trained on the target benchmark dataset.}
\begin{tabular}{c|cc}
\hline
\bf Model &\bf Genre F1 & \bf Style F1 \\ \hline
melody2vec & 0.649 & 0.299 \\ 
tonnetz & 0.627 & 0.253 \\
pianoroll & 0.640 & 0.365 \\
PiRhDy  & 0.668 & 0.471 \\ \
MusicBERT  &  0.784 & 0.645 \\ \hline
Chance & 0.077 & 0.040\\
LLM  & 0.150 &0.108  \\ 
CNN + LLM-MIDI  & 0.453& 0.352\\ \hline
\end{tabular}
\label{tab:recog}
\end{table}

\begin{table*}[h!]
\caption{Example of LLM's thinking process for classifying genres and styles.}
\centering
\small
\begin{tabular}{c|>{\centering\arraybackslash}p{15cm}}
\hline
\textbf{Prompt} & "What genre is the song described in the following JSON for MIDI file?
\{"melody": [[62, 480, 90, 0], ..." \\ \hline
\textbf LLM &\begin{tabular}{@{}p{15cm}@{}}
"\textbf{1. Melody}\\Notes: Mostly stepwise motion (C, D, E, G, A, B, C, etc.), suggesting a major scale.\\
Rhythm: Consistent quarter-note durations (480 ticks per note).\\
Impression: This kind of movement is common in classical, folk, or simple pop melodies.\\
\textbf{2. Chords}\\
Chord progression: Chord progression: C Major (C-E-G) → F Major (F-A-C) → G Major (G-B-D) → C Major (C-E-G). Standard I-IV-V-I structure, commonly found in pop, classical, and folk music. Chords last for one full beat each, reinforcing harmonic stability.\\
\textbf{3. Bass}\\
Root-note bass at C2 (36 MIDI number), repeating on every beat, which is common in pop, folk, or classical music.\\
\textbf{4. Rhythm Section}\\
Kick (35), snare (38), hi-hat (42) forming a regular backbeat. High repetition rate (240 ticks per hit) suggests a fast tempo. This pattern is common in pop, rock, or even upbeat jazz styles.\\
Given the chord progression, stepwise melody, and backbeat rhythm, this music most likely belongs to:\\
\textbf{Classification:} Pop / Folk – The simple harmonic structure and steady rhythm fit many folk and pop songs."\\
%Classical (Simple Waltz or Children's Song) – %The structure is very traditional, resembling classical beginner pieces.\\
%Soft Rock – If played with a modern band, this could resemble soft rock or singer-songwriter styles. 
\end{tabular}\\ \hline
\end{tabular}
\label{tab:ex}
\vspace{1mm}
\end{table*}

\begin{table*}[t!]
\centering
\caption{Results on melody completion task. Bottom row indicates a transformer model trained with LLM-MIDI dataset. Note that all baseline models are trained on the target benchmark dataset.}
\begin{tabular}{c|ccccc}
\hline
\bf Model & \bf MAP & \bf HITS@1 & \bf HITS@5& \bf HITS@10& \bf HITS@25 \\ \hline
melody2vec & 0.646 &0.578&0.717& 0.774 & 0.867  \\ 
tonnetz & 0.683 & 0.545 &0.865&0.946 & 0.993 \\
pianoroll & 0.762 & 0.645 & 0.916 & 0.967 & 0.995\\
PiRhDy & 0.971 &0.950&0.995& 0.998 & 0.999  \\ \
MusicBERT & 0.985 &0.975&0.997& 0.999 & 1.000  \\ \hline
Chance & 0.078 & 0.020 & 0.100 & 0.200 & 0.500 \\
Transformer + LLM-MIDI &0.153&0.085&0.229 & 0.357 &0.725   \\ \hline
\end{tabular}
\label{tab:melody}
\end{table*}

For melody completion task, we trained a vanilla transformer model of encoder-decoder architecture with 2 layers each, 4 attention heads, and 512-unit feedforward layers, using multi-head self-attention to model temporal dependencies. 128 pitches are linearly mapped to 128-dimensional embedding space. We also trained this model with LLM-MIDI dataset, and use PiRhDy dataset \cite{Liang2020PiRhDyLP} for evaluation, where each test sample consists of 1 positive sample and 49 negative samples, and the goal is to find the correct consecutive phrase for a given melodic phrase. During training, transformer learns to generate plausible continuations that align with the LLM-MIDI dataset’s melodic patterns, leveraging attention to capture short and long-term dependencies (Figure~\ref{fig:melody}). We use teacher forcing by providing the ground-truth target sequence as input to the decoder, predicting each step based on prior true notes rather than its own predicted outputs. During evaluation, each of the 50 candidates is converted to a $64\times128$ binary piano roll, after which we compute a similarity score using cosine similarity between the decoder’s probability matrix and the candidate’s binary matrix for ranking the candidates. For evaluation metrics, we used mean average precision (MAP) to assess ranking quality of predicted pitches and HITS at varying degrees of $k$ to measure the proportion of true pitches in the top $k$ predictions per time step.

Both models were trained for 20 epochs with a batch size of 32, using Adam \cite{Kingma2014AdamAM}. We compare our model's performance to that of melody2vec \cite{Hirai2019Melody2VecDR}, tonnetz \cite{Chuan2018ModelingTT}, pianoroll \cite{dong2018pypianoroll}, PiRhDy \cite{Liang2020PiRhDyLP}, and MusicBERT \cite{Zeng2021MusicBERTSM} for illustrative purpose. Note that all baseline models are trained on target benchmark dataset for respective tasks, whereas our neural network models are trained entirely on LLM-MIDI dataset. We acknowledge that this experimental setup, while demonstrating the capabilities of LLM-generated data, does not provide a direct benchmark against state-of-the-art models trained on real human-composed music. A direct comparison with other highly performant music classification and generation models would require substantial additional experiments, including training our CNN/transformer models on real data or adapting state-of-the-art models to our constrained dataset. This comprehensive benchmarking is a crucial direction for future work to fully situate our findings within the broader landscape. %and cross-entropy loss summed across genre or style outputs. The melody generator was likewise trained for 20 epochs with the same batch size and optimizer, using binary cross-entropy loss to compare generated and original melody tracks. 

\subsection{Results \& Discussion}
\subsubsection{Genre / Style Classification}

Table~\ref{tab:recog} summarizes the results from genre and style classification tasks for both directly asking LLM to classify the input and classifying with CNN trained on LLM-MIDI dataset. The classification results from directly asking LLM to classify the input are higher than chance rate, indicating some ability to discern musical structure. Training CNN on LLM-MIDI dataset outperformed directly asking LLM by a large margin, affirming that supervised training with actual music data provides richer clues for classification than an LLM entirely trained on text. In particular, CNN trained on LLM-MIDI outperformed melody2vec and tonnetz on style classification trained on existing MIDI dataset. Overall, however, our models remained significantly poor compared to supervised models trained on standard music datasets.

For classification task, correctness was determined by a strict string match between the LLM's output and the single genre/style label provided in the TOP-MAGD and MASD taxonomies for each piece. We acknowledge that the LLM is not limited to these vocabularies and that music can often have multiple correct genre/style classifications. However, for a quantifiable initial assessment and consistent comparison with other systems, a strict match was used. This approach has limitations as it may penalize musically reasonable but non-exact classifications. Future work will explore more flexible evaluation metrics, such as using human evaluators to assess musical correctness or employing similarity metrics over genre/style embedding spaces.

Several factors contribute to this outcome. For directly asking LLM, the LLM’s training on text lacks explicit musical context, forcing it to infer genre and style from abstract JSON patterns without prior exposure to MIDI conventions or music theory, unlike models fine-tuned on curated datasets. Also, the zero-shot nature of the task, relying on prompt-based reasoning rather than learned embeddings, inevitably limits precision, as the LLM may prioritize salient string patterns, such as rhythm density, over deeper harmonic or melodic relationships. Additionally, occasional JSON interpretation errors or ambiguous outputs further degrade performance. 

Table~\ref{tab:ex} shows an example of LLM's output for classification task. It shows a structured, analytical approach to interpreting musical attributes from the MIDI JSON data, breaking it down into melody, chords, bass, and rhythm components before synthesizing these observations into genre and style predictions. It identifies melodic motion and scale, chord progression, a steady root-note bass, and a repetitive backbeat rhythm, and map these to genres. This suggests that LLM leverages its text-based training to recognize patterns akin to music theory concepts, such as harmonic stability and rhythmic consistency, demonstrating its capability for zero-shot music analysis without explicit musical pre-training. Its ability to cross-reference features across tracks and propose multiple plausible classifications highlights a nuanced pattern-matching ability of musical conventions, offering a scalable, prompt-driven alternative to traditional models. However, limitations emerge in its lack of precision and depth: the analysis relies on observable string patterns without considering subtler cues like timbre, or long-term structure, which are absent in current text input and critical for distinguishing nuanced styles. It is important to clarify that MIDI velocity, which encodes dynamics, was indeed part of the JSON input, but its textual reasoning seems to have prioritized other features. It also depends highly on generalized knowledge for classification, rather than domain-specific musical expertise.

For CNN trained with LLM-MIDI dataset, the fixed multi-track structure may have oversimplified the musical features, missing nuances like polyphony or dynamic instrumentation critical for accurate classification, especially in complex genres like \textit{jazz} or styles like \textit{minimalist}. While surpassing chance rate suggests the LLM captures rudimentary musical cues, its poor overall performance highlights a gap in translating string-based knowledge to the structured, temporal domain of music, underscoring the need for musical grounding or enhanced prompting strategies to rival specialized systems. Moreover, the fact that a CNN trained on LLM-generated MIDI outperforms the direct LLM classification further supports the notion that LLMs primarily model token sequences. While the LLM is adept at generating sequences consistent with textual descriptions, its ``understanding" is not sufficient for high-accuracy direct classification when compared to models specifically trained on musical data. This suggests that the observed capabilities are better explained by sophisticated pattern matching rather than a deep, human-like musical comprehension.

\subsubsection{Melody Completion}

Table~\ref{tab:melody} summarizes the results from melody completion task. 
Our transformer’s melody completion performance also exceeds random chance throughout all evaluation metrics, suggesting it was able to learn melodic structure from the LLM-generated dataset to some extent, capturing basic patterns such as pitch continuity or rhythmic flow. However, as with the classification tasks, it still falls far below supervised models trained with existing music datasets, which leverage human-composed data to achieve near-perfect rankings. This gap underscores the challenges of relying on synthetic, LLM-derived training data for a task requiring precise melodic coherence.

Our transformer model’s modest success reflects its ability to generalize short-term dependencies from the LLM-MIDI dataset, yet several task-specific factors limit its effectiveness. The encoder-decoder architecture, while adept at modeling local note transitions, struggles to align the synthetic dataset’s often erratic continuations with human-crafted candidates, leading to lower similarity scores and poorer rankings. Unlike classification, where the CNN could exploit multi-track features, the melody completion task isolates the melody track, amplifying the impact of the LLM’s inconsistent outputs, without contextual support from chords or rhythm. Nevertheless, the above-chance performance highlights potential: with refinements such as training on longer sequences or incorporating musically informed similarity measures, training a transformer model on LLM-generated dataset could potentially further narrow down the gap to established systems, leveraging the LLM’s scalability for creative melody completion applications.

%In conclusion, our results highlight both the potential and limitations of LLMs’ internal perception of symbolic music. The ability to generate classifiable structures and plausible melodies from text alone underscores a remarkable capacity for cross-domain pattern recognition, suggesting that LLMs can infer a rudimentary musical world from string modeling. However, their performance lags behind specialized systems due to output inconsistency, structural rigidity, and lack of musical grounding. These findings affirm LLMs as promising tools for creative synthesis while emphasizing the need for domain-specific enhancements to fully unlock their musical potential.

\section{Conclusion}
\label{sec:conclusion}

In this paper, we explored the extent to which modeling relationships between strings teaches large language models about the musical world and how LLMs implicitly model symbolic music, by generating a large-scale dataset of symbolic music and evaluating its utility in classification and generation tasks. We demonstrated that LLMs can generate musically structured outputs, comprising melody, chords, bass, and rhythm tracks, without explicit musical training. Our experiments also revealed that the LLM-generated music encodes distinguishable features of genre and style to some degree, as evidenced by the performance of CNN trained on LLM-generated dataset, achieving better results than chance rate. We also showed that it is possible to learn melody patterns by training a neural network entirely on LLM-generated dataset. While the performance is considerably lower than models trained on human-crafted music dataset for all tasks, which is an easily predictable result that aligns with intuition, note that our goal is not to outperform baseline models, but to highlight both the limitations and potential of LLMs in musical contexts. The performance gap to supervised benchmarks underscores the need for musical grounding or refined generation strategies, but the ability to exceed chance rate suggests an inherent capacity to capture rudimentary musical structures through text-based learning. As such, the contribution of our work lies not in immediate practicality but in its exploratory insight, as it reveals LLMs’ latent capacity to bridge text and symbolic music for cross-domain learning without domain-specific data. %This positions our study as a contribution to understanding AI’s representational boundaries, inspiring novel research directions in music informatics despite its current limitations.

In conclusion, this paper illuminates how LLMs encode musical knowledge through text, offering a novel perspective into their representational power and paving the way for innovative generative music applications. The implications of the results can be important for both AI research and music informatics, as it positions LLMs as versatile learners capable of cross-domain synthesis, reinforcing the notion that string modeling fosters generalized understanding of structured systems.

\bibliography{aaai2026}

\end{document}